\documentclass[11pt,a4paper]{article}
\usepackage[hyperref]{emnlp-ijcnlp-2019}
\usepackage{times}
\usepackage{latexsym}
\usepackage{times}
\usepackage{latexsym}
\usepackage{amsfonts,amssymb}
\usepackage{graphicx}
\usepackage{stfloats}
\usepackage{multirow}
\usepackage{verbatim}

\usepackage{url}

\aclfinalcopy

\title{Subword ELMo}

\author{Jiangtong Li$^{1,}$\thanks{$^{*}$Work done when Jiangtong Li was interning at Tencent AI Lab. $^{\dag}$ Corresponding author. This paper was partially supported by National Key Research and Development Program of China (No. 2017YFB0304100) and Key Projects of National Natural Science Foundation of China (No. U1836222 and No. 61733011).} ,  Hai Zhao$^{1,2,3,\dag}$, Zuchao Li$^{1,2,3}$, Wei bi $^{4}$, Xiaojiang Liu $^{4}$  \\
 $^{1}$Department of Computer Science and Engineering, Shanghai Jiao Tong University \\
 $^{2}$Key Laboratory of Shanghai Education Commission for Intelligent Interaction \\ and Cognitive Engineering, Shanghai Jiao Tong University, Shanghai, China\\
 $^{3}$MoE Key Lab of Artificial Intelligence, AI Institute, Shanghai Jiao Tong University \\
 $^4$Tencent AI Lab, Shenzhen, China\\
 \tt jiangtongli1997@gmail.com, zhaohai@cs.sjtu.edu.cn,  \\ \tt charlee@sjtu.edu.cn, \{victoriabi, kieranliu\}@tencent.com
}

\date{}

\begin{document}

\maketitle

\begin{abstract}
 Embedding from Language Models (ELMo) has shown to be effective for improving many natural language processing (NLP) tasks, and ELMo takes character information to compose word representation to train language models. 
 However, the character is an insufficient and unnatural linguistic unit for word representation. 
 Thus we introduce Embedding from Subword-aware Language Models (ESuLMo) which learns word representation from subwords using unsupervised segmentation over words.
 We show that ESuLMo can enhance four benchmark NLP tasks more effectively than ELMo, including syntactic dependency parsing, semantic role labeling, implicit discourse relation recognition and textual entailment, which brings a meaningful improvement over ELMo.
\end{abstract}

\section{Introduction}

Recently, pre-trained language representation has shown to be useful for improving many NLP tasks \cite{peters2018deep, devlin2018bert, radford2018improving, howard2018universal}. 
Embeddings from Language Models (ELMo) \cite{peters2018deep} is one of the most outstanding works, which uses a character-aware language model to augment word representation.

An essential challenge in training word-based language models is how to control vocabulary size for better rare word representation. 
No matter how large the vocabulary is, rare words are always insufficiently trained. Besides, an extensive vocabulary takes too much time and computational resource for the model to converge. 
Whereas, if the vocabulary is too small, the out-of-vocabulary (OOV) issue will harm the model performance heavily \cite{jozefowicz2016exploring}.
To obtain effective word representation,  \citet{jozefowicz2016exploring} introduce character-driven word embedding using convolutional neural network (CNN) \cite{lecun1998gradient}, following the language model in \citet{kim2016character} for deep contextual representation.

However, there is potential insufficiency when modeling word from characters which hold little linguistic sense, especially, the morphological source \cite{fromkin2018introduction}. 
Only 86 characters(also included some common punctuations) are adopted in English writing, making the input too coarse for embedding learning. 
As we argue that for better representation from a refined granularity, word is too large and character is too small, it is natural for us to consider subword unit between character and word levels.

Splitting a word into subwords and using them to augment the word representation may recover the latent syntactic or semantic information \cite{bojanowski2017enriching}.
For example, \textit{uselessness} could be split into the following subwords: $<$\textit{use}, \textit{less}, \textit{ness}$>$. 
Previous work usually considers linguistic knowledge-based methods to tokenize each word into subwords (namely, morphemes) \cite{luong2010hybrid,luong2013better,bhatia2016morphological}.
However, such treatment may encounter three main inconveniences. 
First, the subwords from linguistic knowledge, typically including the morphological suffix, prefix, and stem, may not be suitable for a targeted NLP task \cite{banerjee2018meaningless} or mislead the representation of some words, like the meaning of \textit{understand} cannot be formed by \textit{under} and \textit{stand}.
Second, linguistic knowledge, including related annotated lexicons or corpora, may not even be available for a specific low-resource language.
Due to these limitations, we focus on computationally motivated subword tokenization approaches in this work.


In this paper, we propose Embedding from Subword-aware Language Models (ESuLMo), which takes subword as input to augment word representation and release a sizeable pre-trained language model research communities.
Evaluations show that the pre-trained language models of ESuLMo outperform all RNN-based language models, including ELMo, in terms of PPL and ESuLMo outperforms state-of-the-art results in three of four downstream NLP tasks.



\begin{figure}[t]
\centering
\includegraphics[scale=0.47]{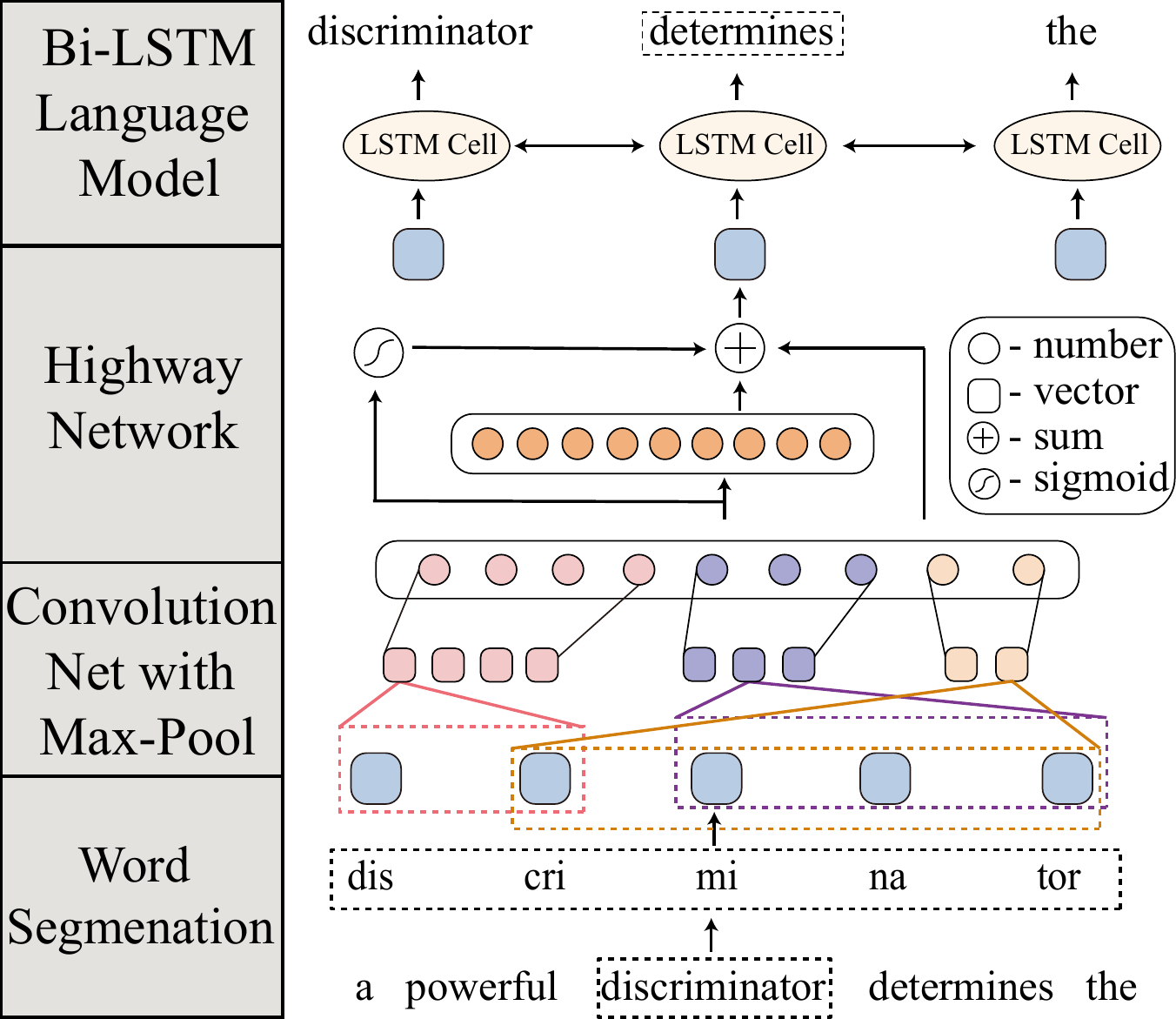}
\caption{Model Structure (We replace the input block by the Word Segmentation block.)}
\label{fig:main}
\end{figure}

\section{General Language Model}\label{LM}

The overall architecture of our subword-aware language model shows in Figure \ref{fig:main}. It consists of four parts, word segmentation, word-level CNN, highway network and sentence-level RNN.

Given a sentence $S = \{W_1, W_2, ... , W_n\}$, we first use a segmentation algorithm to divide each word into a sequence of subwords \cite{sennrich2015neural,kudo2018subword}.
\begin{eqnarray}
M_i = \{x_{i, 1}, x_{i, 2}, ... ,x_{i, m}\} = f(X_i)
\end{eqnarray}
where $M_i$ is the output of the segmentation algorithm, $x_{i, j}$ is the subword unit and $f$ represents the segmentation algorithm. Then a look-up table is applied to transform the subword sequence into subword embeddings \cite{mikolov2013efficient}. 

To further augment the word representation from the subwords, we apply a narrow convolution between subword embeddings and several kernels.
\begin{eqnarray}
\mathbf{y}_i = Concat(g(\mathbf{M}_i, \textbf{K}_1), g(\mathbf{M}_i, \textbf{K}_2), ...), \label{eq1}
\end{eqnarray}
where $Concat$ is the concatenation operation for all the input vectors, $\mathbf{K}_i$ is convolution kernel and $g$ is CNN-MaxPooling operation.

A highway network \cite{srivastava2015training} is then applied to the output of CNN. 
A bidirectional long short-term memory network (Bi-LSTM)  \cite{hochreiter1997long} generates the hidden states for the given sentence representation in forward and backward.
Finally, the probability of each token is calculated by applying an affine transformation to all the hidden states followed by a $SoftMax$ function.
During the training, our objective is to minimize the negative log-likelihood of all training samples.

To apply our pre-trained language models to other NLP tasks, we combine the input vector and the last layer's hidden state of the Bi-LSTM to represent each word.

\section{Subword from Unsupervised Segmentation}\label{inputrep}

To segment subwords from a word, we adopt the generalized unsupervised segmentation framework proposed by \citet{zhao2008seg}. 
The generalized framework can be divided into two collocative parts, goodness measure (score), which evaluates how likely a subword is to be a ‘proper’ one, and a segmentation or decoding algorithm. 
For the sake of simplicity, we choose frequency as the goodness score and two representative decoding algorithms, byte pair encoding (BPE) ~\cite{sennrich2015neural} which uses a greedy decoding algorithm and unigram language model (ULM) ~\cite{kudo2018subword} which adopts a Viterbi-style decoding algorithm.

For a group of character sequences, the working procedure of BPE is as follows:

\noindent $\bullet$ All the input sequences are tokenized into a sequence of single-character subwords.

\noindent $\bullet$ Repeatedly, we calculate the frequencies of all bigrams and merge the bigram with the highest one until we get the desired subword vocabulary.

ULM is proposed based on the assumption that each subword occurs independently.
The working procedure of ULM segmentation is as follows.

\noindent $\bullet$ Heuristically make a reasonably large seed vocabulary from the training corpus.

\noindent $\bullet$ Iteratively, the probability of each subword is estimated by the expectation maximization (EM) algorithm and the top $\eta\%$ subwords with the highest probabilities are kept. 
Note that we always keep the single character in subword vocabulary to avoid out-of-vocabulary.

For a specific dataset, the BPE algorithm keeps the same segmentation for the same word in different sequences, whereas ULM cannot promise such segmentation. 
Both segmentation algorithms have their strengths, \citet{sennrich2015neural} show that BPE can fix the OOV issue well, and \citet{kudo2018subword} proves that ULM is a subword regularization which is helpful in neural machine translation.

\section{Experiments}

The ESuLMo is evaluated in two ways, task independent and task dependent.
For the former, we examine the perplexity of the pre-trained language models.
For the latter, we examine on four benchmark NLP tasks, syntactic dependency parsing, semantic role labeling, implicity discourse relation recognition, and textual entailment.

\subsection{Language Model}

\begin{table}[h]
\centering
\small
\setlength{\tabcolsep}{1mm}{
\begin{tabular}{lllll}
\hline
\multicolumn{3}{l}{Model} & PPL & \#Params \\ \hline
\multicolumn{2}{l}{BIG G-LSTM-2} & \shortcite{Oleksii2017FACTORIZATION} & 36 & - \\
BIG LSTM & Char & \shortcite{jozefowicz2016exploring} & 30.0 & 1.8B \\
ELMo\footnote{The result of ELMo is re-implemented by ourselves.} & Char & \shortcite{peters2018deep} & 29.3(39.9) & 1.94B \\ \hline
\multirow{6}{*}{ESuLMo} & \multirow{6}{*}{Sub} & BPE,500 & \textbf{27.6(40.3)} & 1.95B \\
 &  & BPE,1000 & 28.1(42.9) & 1.96B \\
 &  & BPE,2000 & 28.6(44.4) & 1.96B \\
 &  & ULM,500 & 28.9(43.8) & 1.95B \\
 &  & ULM,1000 & 30.7(44.1) & 1.96B \\
 &  & ULM,2000 & 31.5(50.4) & 1.96B \\ \hline
\end{tabular}}
\caption{The converged\footnote{If the PPL decay between two epochs is less than one, the model is regarded as converged.} test PPL (numbers in the parentheses are the PPL after 10 epochs).}
\label{tb:up}
\end{table}

\noindent In this section, we examine the pre-trained language models of ESuLMo in terms of PPL.
All the models' training and evaluation are done on One Billion Word dataset \cite{chelba2013one} \footnote{Our code and models were released publicly in \href{https://github.com/Jiangtong-Li/Subword-ELMo/}{https://github.com/Jiangtong-Li/Subword-ELMo/}}.
During training, we strictly follow the same hyper-parameter published by ELMo, including the hidden size, embedding size, and the number of LSTM layers. Meanwhile, we train each model on 4 Nvidia P40 GPUs, which takes about three days for each epoch. 
Table \ref{tb:up} shows that our pre-trained language models can improve the performance of RNN-based language models by a large margin and our subword-aware language models outperform all previous RNN-based language models, including ELMo, in terms of PPL. 
During the experiment, we find that 500 is the best vocabulary size for both segmentation algorithms, and BPE is better than ULM in our setting.

\subsection{Downstream Tasks}\label{exp:d}

\begin{table}[]
\scriptsize
\setlength{\tabcolsep}{0.5mm}{
\begin{tabular}{ll|llll}
\hline \hline
\multicolumn{2}{l|}{Tasks}         & SDP & SRL & IDRR & TE \\ \hline 
\multicolumn{2}{l|}{SOTA(Single Model)}  & 96.35 \shortcite{wang2018improved}  & 90.4 \shortcite{li2019srl}  & 48.22 \shortcite{bai2018deep} & 91.1 \shortcite{liu2019multi}   \\ \hline
\multicolumn{2}{l|}{Our Baseline}  & 95.83 \shortcite{dozat2016deep}     & 89.6 \shortcite{cai2018srl} & 47.03 \shortcite{bai2018deep} & 88.0 \shortcite{chen2016enhanced} \\ \hline
ELMo & Char(86)          & 96.45 & 90.0 & 48.22 & 88.7 \\ \hline
\multirow{6}{*}{ESuLMo} &BPE(500)      & \textbf{96.65$^{+}$} & \textbf{90.5$^{+}$} & 48.99$^{+}$ & \textbf{89.5$^{+}$} \\
&BPE(1000)     & 96.62$^{+}$ & 90.4$^{+}$ & \textbf{49.07$^{+}$} & 89.4$^{+}$ \\
&BPE(2000)     & 96.54  & 90.4$^{+}$ & 48.88$^{+}$ & 89.2$^{+}$ \\
&ULM(500)      & 96.55 & 90.2$^{+}$ & 48.73$^{+}$ & 89.1$^{+}$ \\
&ULM(1000)     & 96.51 & 90.0 & 48.32 & 88.9 \\
&ULM(2000)     & 96.44 & 90.0 & 48.35 & 88.7 \\ \hline \hline
\end{tabular}}
\caption{ELMo vs. ESuLMo on four NLP tasks. $^{+}$ indicates the performance compared to that of ELMo is statistically significant.}
\label{tb:down}
\end{table}

While applying our pre-trained ESuLMo to other NLP tasks, we have two different strategies: 
(1) Fine-tuning our ESuLMo while training other NLP tasks; 
(2) Fixing our ESuLMo while training other NLP tasks. 
During the experiment, we find there is no significant difference between these two strategies.
However, the first strategy consumes much more resource than the second one. 
Therefore, we choose the second strategy to conduct all the remaining experiments.

We apply ESuLMo to four benchmark NLP tasks. And we choose the fine-tuned model by validation set and report the results in the test set.
The comparisons in Table \ref{tb:down} show that ESuLMo outperforms ELMo significantly in all tasks and achieves the new state-of-the-art result in three of four tasks \footnote{During the experiment, we use the ESuLMo in the same way as in the ELMo. 
All result in Sec \ref{exp:d} use the subword-aware language model trained for 10 epochs (same as ELMo), and each result is the average across five random seed.}.

\textbf{Syntactic Dependency Parsing (SDP)} is to disclose the dependency structure over a given sentence. \citet{dozat2016deep} use a Bi-LSTM encoder and a bi-affine scorer to determine the relationship between two words in a sentence. 
Our ESuLMo gets 96.65\% UAS in PTB-SD 3.5.0, which is better than the state-of-the-art result \cite{wang2018improved}.

\textbf{Semantic Role Labeling (SRL)} is to model the predicate-argument structure of a sentence. 
\citet{cai2018srl} model SRL as a words pair classification problem and directly use a bi-affine scorer to predict the relation given two words in a sentence. 
By adding our ESuLMo to the baseline model \cite{cai2018srl}, we can not only outperform the original ELMo by 0.5\% F1-score but also outperform the state-of-the-art model \cite{li2019srl} which has three times more parameters than our model in CoNLL 2009 benchmark dataset.

\textbf{Implicit Discourse Relation Recognition (IDRR)} is a task to model the relation between two sentences without explicit connective.
\citet{bai2018deep} use a hierarchical structure to capture four levels of information, including character, word, sentence and pair. We choose it as our baseline model for 11-way classification on PDTB 2.0 following \citet{ji2015one}'s setting. 
Our model outperforms ELMo significantly and reaches a new state-of-the-art result.

\textbf{Textual Entailment (TE)} is a task to determine the relationship between a hypothesis and a premise. 
The Stanford Natural Language Inference (SNLI) corpus \cite{bowman2015large} provides approximately 550K hypothesis/premise pairs. 
Our baseline adopts ESIM \cite{chen2016enhanced} which uses a Bi-LSTM encoder layer and a Bi-LSTM inference composition layer which are connected by an attention layer to model the relation between hypothesis and premise. 
Our ESuLMo outperforms ELMo by 0.8\% in terms of accuracy. Though our performance does not reach the state-of-the-art, it is second-best in all single models according to the SNLI leaderboard \footnote{https://nlp.stanford.edu/projects/snli/. The second-best claim is based on the leaderboard when this work is submitted to EMNLP 2019.}.

\section{Discussion}

\noindent\textbf{Subword Vocabulary Size}
Tables \ref{tb:up} and \ref{tb:down} show the performance of ESuLMo drops with the vocabulary size increases \footnote{Considering that the original ELMo is also a kind of ESuLMo with the vocabulary size of 86. We have examined ESuLMo with 250 subwords, and no significant performance difference was observed, and thus, the corresponding results are not reported here due to the limited space.}. 
We explain the trend that neural network pipeline especially CNN would fail to capture
necessary details of building word embeddings
as more subwords are introduced.

\noindent\textbf{Subword Segmentation Algorithms}\label{segalg}
Tables \ref{tb:up} and \ref{tb:down} show that ESuLMo based on both ULM and BPE segmentation with 500 subwords outperform the original ELMo, and BPE is consistently better than ULM on all evaluations under the same settings.
We notice that BPE can give static subword segmentation for the same word in different sentences, while ULM cannot. It suggests that ESuLMo is sensitive to segmentation consistency.

We also analyze the subword vocabularies from two algorithms and find that the overlapping rates for 500, 1K and 2K sizes are 60.2\%, 55.1\% and 51.9\% respectively. This indicates subword mechanism can stably work in different vocabularies\footnote{Two full vocabularies of size 500 are in Appendix.}.

\begin{figure}[h]
\centering
\includegraphics[scale=0.28]{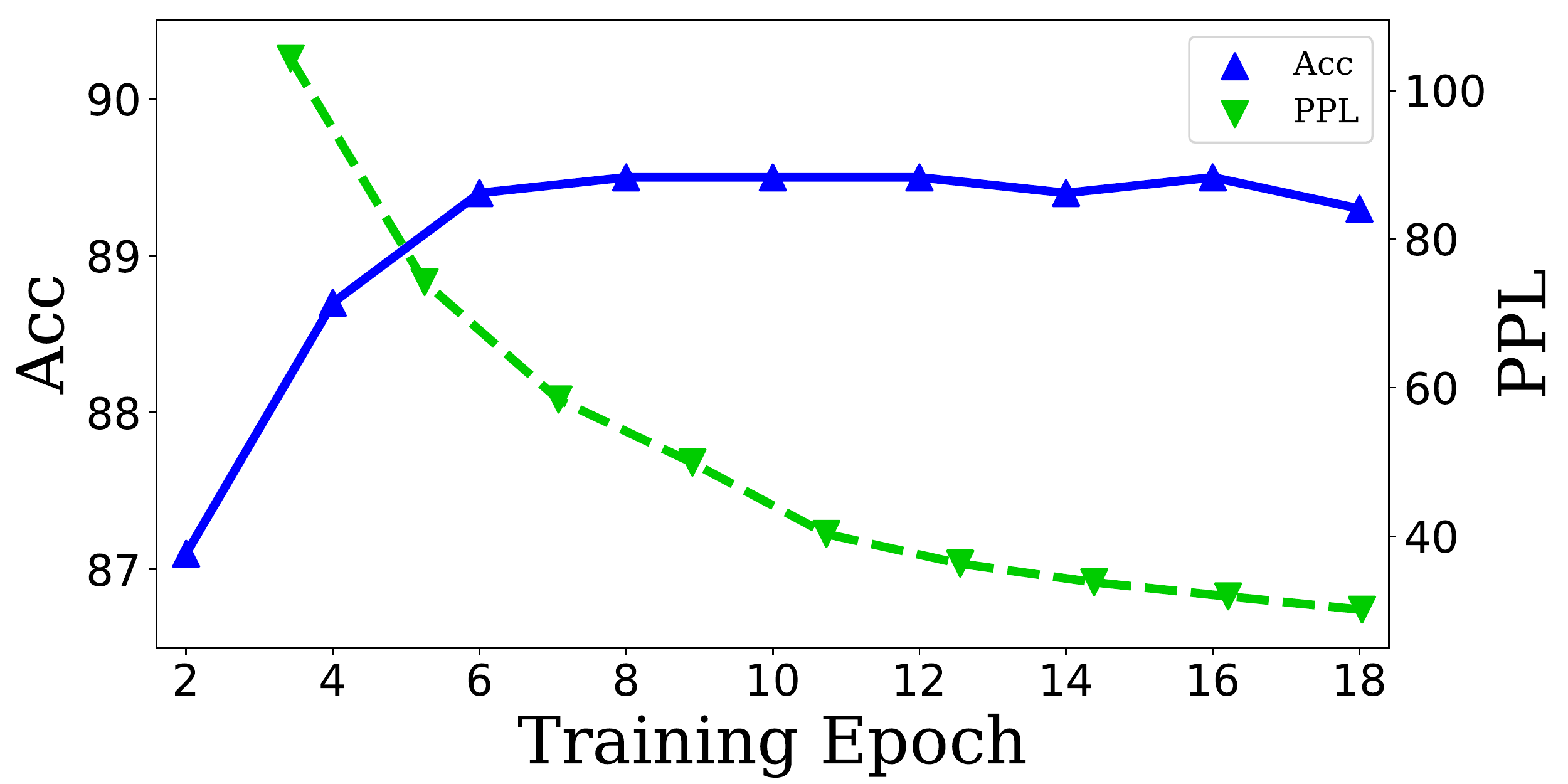}
\caption{Accuracy in SNLI vs. PPL for different training epochs. (For SNLI, the Training Epoch means that how many epochs the language model are trained before we use it.)}
\label{fig:train}
\end{figure}

\noindent\textbf{Task Independent vs. Task Specific}
To discover the necessary training progress, we show the accuracy in SNLI and PPL for language model in Figure \ref{fig:train}. 
The training curves show that our ESuLMo helps ESIM reach stable accuracy for SNLI while the corresponding PPL of the language model is far away from convergence.

\begin{table}[ht]
\centering
\small
\begin{tabular}{ll}
\hline \hline
Model                      & F1 score \\ \hline\hline
WordNet 1st Sense Baseline & 65.9     \\
\citet{alessandro2017neural}& 69.9     \\
\citet{ignacio2016embeddings}                    & \textbf{70.1}     \\ \hline
ELMo                       & 69.0     \\ 
ESuLMo                     & 69.6        \\ \hline\hline
\end{tabular}
\caption{Word Sense Disambiguation}
\label{tb:wsd}
\end{table}

\noindent\textbf{Word Sense Disambiguation}
To explore the word sense disambiguation capability of our ESuLMo, we isolate the representation encoded by our ESuLMo and use them to directly make predictions for a fine-grained word sense disambiguation (WSD) task\footnote{A nearest neighbor disambiguation example is in Appendix.}.
We choose the dataset and perform this experiment using the same setting as ELMo with only the last layer's representation. 
Table \ref{tb:wsd} shows that our model can outperform the original ELMo.

\section{Conclusion}
In this paper, we present Embedding from Subword-aware Language Model (ESuLMo).
The experiments show that the language models of ESuLMo outperform all RNN-based language models, including ELMo, in terms of PPL. 
The empirical evaluations in benchmark NLP tasks show that subwords can represent word better than characters to let ESuLMo more effectively promote downstream tasks than the original ELMo.

\bibliography{emnlp-ijcnlp-2019}
\bibliographystyle{acl_natbib}

\end{document}